\documentclass[conference,a4paper]{IEEEtran}
\IEEEoverridecommandlockouts

\usepackage{makecell}
\usepackage{graphicx}
\usepackage{xcolor}

\usepackage{dblfloatfix}

\usepackage[hidelinks]{hyperref}
\usepackage[cmex10]{amsmath}
\usepackage{amssymb,amsfonts}
\usepackage{multirow}
\usepackage{booktabs}
\usepackage{graphicx}
\usepackage{rotating}

\usepackage[numbers,compress]{natbib}

\usepackage{booktabs}
\usepackage{multirow}
\usepackage{graphicx}
\usepackage{rotating}

\usepackage[ruled,vlined]{algorithm2e}
\usepackage{graphicx}
\usepackage{svg}
\graphicspath{{Figures/PDF/}{Figures/PNG/}}
\usepackage{balance}

\usepackage{booktabs}
\usepackage{siunitx}
\usepackage{texnames}
\usepackage{bm,bbm}
\usepackage{orcidlink}

\usepackage{filecontents} 

\usepackage[absolute,overlay]{textpos}
\usepackage{ragged2e}
\newcommand{\ieeecopyrightnotice}{%
  \begin{textblock*}{195mm}(7.5mm,8mm)
    \noindent{\footnotesize\justifying\copyright~2026 IEEE. Personal use of this material is permitted. Permission from IEEE must be obtained for all other uses, in any current or future media, including reprinting/republishing this material for advertising or promotional purposes, creating new collective works, for resale or redistribution to servers or lists, or reuse of any copyrighted component of this work in other works.\par}
  \end{textblock*}}

\begin{document}
\bstctlcite{IEEEexample:BSTcontrol}

\title{\uppercase{Methane-plume segmentation from hyperspectral satellite imagery via  multimodal deep learning 
}
}

\author{
\IEEEauthorblockN{
Brayan Quintero\textsuperscript{\orcidlink{0009-0001-5351-8888}},
Jeferson Acevedo\textsuperscript{\orcidlink{0009-0008-0840-1972}},
Samuel Traslaviña\textsuperscript{\orcidlink{0009-0002-7423-8751}},
Hoover Rueda-Chacón\textsuperscript{\orcidlink{0000-0002-6763-8629}}
}
\IEEEauthorblockA{
\textit{Department of Computer Science, Universidad Industrial de Santander}\\
Bucaramanga, 680002, Colombia\\
\{brayan2221707, jeferson2221790, samuel2220085\}@correo.uis.edu.co, hfarueda@uis.edu.co
}
}

\maketitle
\ieeecopyrightnotice
\begin{abstract}
Efficient detection of methane plumes is crucial for understanding and mitigating global warming, as accurately identifying and segmenting them in earth observation imagery remain essential for large-scale monitoring. In this work, we propose a multimodal deep learning model that integrates a feature-guided methane enhancement (FGME) mechanism which injects physically meaningful methane cues into transformer-based RGB representations at multiple semantic scales. Our method is evaluated on the MPDataset, where it outperforms the state-of-the-art with improvements of +0.92 in MIoU, +0.87 in MPrecision and +1.01 in Recall. Notably, these gains are obtained with a substantially lower computational cost than other high-performing architectures, resulting in a favorable accuracy–efficiency trade-off for large-scale methane monitoring. These results highlight the potential of efficient multimodal fusion strategies for accurate and scalable methane plume segmentation in real-world remote sensing applications.


\end{abstract}


\section{Introduction}

Methane (CH$_4$) is the second most significant greenhouse gas after carbon dioxide (CO$_2$) \cite{FILONCHYK2024173359}. Despite its relatively low atmospheric abundance, methane exhibits a substantially higher short-term global warming potential. Methane emissions originate from both natural processes, such as wetlands and marine sediments, and anthropogenic activities, including agriculture, fossil fuel extraction and transport, and waste management \cite{essd-17-1873-2025}.

Accurate detection and monitoring of methane emissions remain challenging due to their spatial sparsity, temporal variability, and low signal-to-noise ratio \cite{Jacob2016}. Hyperspectral remote sensing has emerged as a critical technology for large-scale methane observation, as its high spectral resolution enables the detection and delineation of methane plumes from satellite and airborne platforms \cite{THORPE2016104, Kumar_2020_WACV}.

Several deep learning–based approaches have been proposed for methane plume detection by operating directly on gas-derived enhancement representations. For instance, Radman et al. \cite{RADMAN2023113708} generated a benchmark dataset by embedding simulated methane plumes into real Sentinel-2 backgrounds and introduced an end-to-end deep learning framework trained on methane enhancement imagery. Jongaramrungruang et al. \cite{JONGARAMRUNGRUANG2022112809} proposed MethaNet, a convolutional neural network (CNN) capable of directly predicting methane point-source emissions from two-dimensional plume enhancement maps without relying on auxiliary measurements such as wind information. Joyce et al. \cite{Joyce2023} employed a UNet-based architecture for methane plume mask detection in PRISMA (Hyperspectral Precursor of the Application Mission) imagery, incorporating an additional $1\times1$ convolutional layer to emphasize methane-related anomalies.
\begin{figure}[!t]
    \centering
    \includegraphics[width=\linewidth]{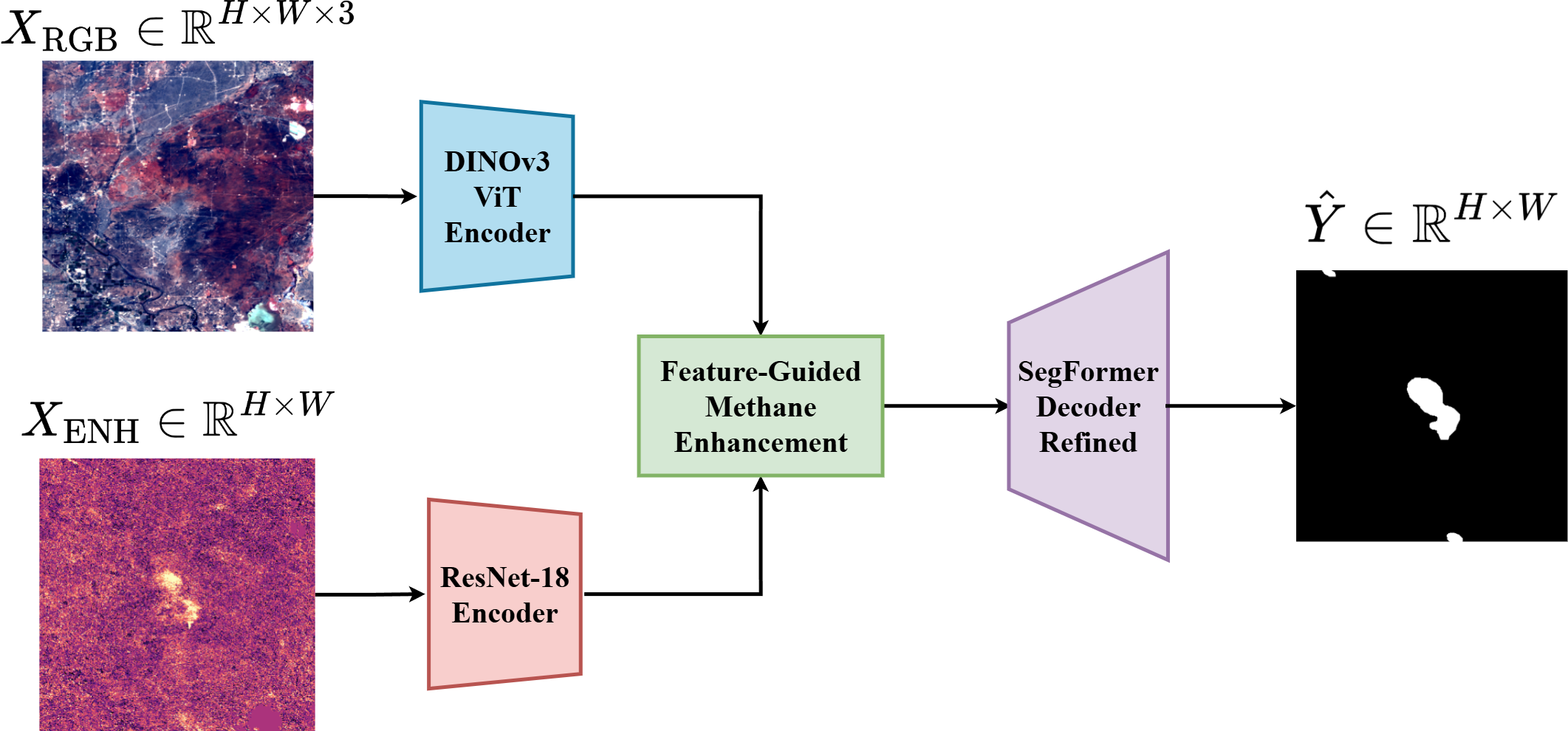}
    \caption{Overview of the proposed framework. RGB imagery $X_{\mathrm{RGB}} \in \mathbb{R}^{H\times W\times 3}$ and methane enhancement map $X_{\mathrm{ENH}} \in \mathbb{R}^{H\times W}$ are processed by modality-specific encoders and fused through a feature-guided methane enhancement (FGME) module. The resulting representations are decoded to produce the final binary methane plume mask $ \hat{Y} \in \mathbb{R}^{H\times W} $.  }
    \label{fig:Framework}
\end{figure}
In this study, we propose a multimodal methane plume segmentation model, where the key intuition is that a physically plausible methane plume should be supported not only by a gas-related spectral signal but also by a coherent spatial structure and scene context. While methane enhancement provides direct evidence of gas presence, RGB imagery offers complementary information related to terrain, infrastructure, and plume morphology. Motivated by this observation, our approach jointly leverages RGB information and methane-guided cues to improve segmentation robustness, producing more physically consistent methane plume detections than state-of-the-art  enhancement-only methods.

\section{Methodology}

\subsection{Overall Framework}

We propose a multimodal deep learning architecture for methane plume segmentation in hyperspectral remote sensing imagery. The model combines morphological RGB data with gas-related information from a methane enhancement map. Let the input be a stacked tensor
\begin{equation}
X = Cat([X_{\mathrm{RGB}}, X_{\mathrm{ENH}}]) \in \mathbb{R}^{H\times W\times 4}, 
\end{equation} 
where “Cat” denotes concatenation along the channel dimension. 

As shown in Fig.~\ref{fig:Framework}, each modality is encoded by a dedicated backbone to extract hierarchical representations. The resulting features are then fused through a methane-guided enhancement module that injects methane enhancement-driven cues into the RGB feature to emphasize plume-consistent patterns. A lightweight segmentation decoder predicts a binary plume mask $ \hat{Y} \in \mathbb{R}^{H\times W} $, which denotes the background and plume classes.


\subsection{Model Architecture}

\subsubsection{\textbf{RGB semantic encoder}}

The RGB $X_{\mathrm{RGB}} \in \mathbb{R}^{H\times W\times 3}$ branch employs a DINOv3 ViT-S/16 \cite{simeoni2025dinov3} backbone to extract semantically rich representations:
\begin{equation}
\{H_{l}\}_{l=1}^{4} = f_{\mathrm{DINO}}(X_{\mathrm{RGB}}).
\end{equation}
We extract four intermediate feature maps from DINOv3 corresponding to transformer blocks 2,5,8 and 11, and concatenate them along the channel dimension to form a tap tensor $T = \mathrm{Cat}(H_{1},\dots,H_{4})$.
A four-level pyramid is then built by progressive average pooling and \(1\times1\) projections:
\begin{equation}
D_l = \mathrm{Conv}^{1 \times 1}_{\text{RGB},l}(\mathrm{Pool}_{2^{l-1}}(T)), \quad l\in\{1,2,3,4\},
\end{equation}
where \(\mathrm{Pool}_{1}\) is the identity and \(\mathrm{Conv}^{1\times1}_{\text{RGB},l}\) denotes a \(1\times1\) convolution that matches the SegFormer decoder channel sizes.

\subsubsection{\textbf{Methane enhancement encoder}}

The methane enhancement channel ${X}_{\mathrm{ENH}} \in \mathbb{R}^{H\times W}$ is processed using a lightweight ResNet-18 \cite{he2016deep} encoder, producing a hierarchy of feature maps:
\begin{equation}
\{M_{l}\}_{l=1}^{4} = f_{\mathrm{ResNet}}(X_{\mathrm{ENH}}),
\end{equation}
which preserve structural and spatial information relevant for localizing methane plumes.

\begin{figure*}[!t]
    \centering
    \includegraphics[width=1.0\linewidth]{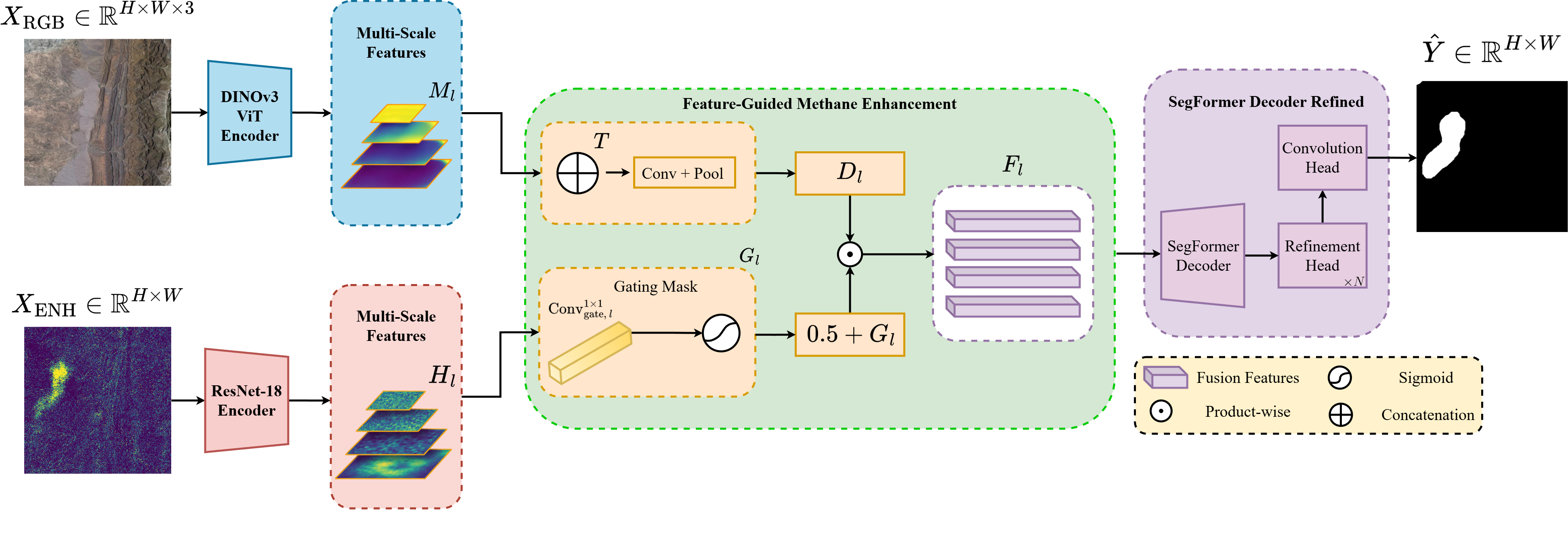}
    \caption{Multimodal architecture for methane plume segmentation. RGB images and methane enhancement maps are processed by DINOv3 and ResNet-18 encoders, respectively, producing multi-scale features. A Feature-Guided Methane Enhancement module adaptively modulates and fuses both modalities, and the resulting representations are decoded by a SegFormer-based head to generate the final plume mask.
    The decoder follows the SegFormer-B1 configuration, initialized from the \texttt{nvidia/segformer-b1-finetuned-ade-512-512} model.
}
    \label{fig:Pipeline}
\end{figure*}

\subsubsection{\textbf{Feature-guided methane enhancement}}

Multimodal fusion is performed through a feature-guided methane
enhancement (FGME) mechanism. At each semantic scale \( l \in \{1,\dots,4\} \), the corresponding RGB and methane feature maps \(D_l\) and \(M_l\) are fused to produce a guided feature representation \(F_l\). Prior to gating, \(M_l\) is bilinearly resized when necessary to match the spatial resolution of \(D_l\). The gating mask is computed from methane features using a \(1\times1\) projection followed by a sigmoid activation:

\begin{equation}
G_l = \sigma(\mathrm{Conv}^{1 \times 1}_{\text{gate},l}(M_l)).
\end{equation}
The fused feature map is then obtained as:

\begin{equation}
F_{l} = D_{l} \odot (0.5 + G_{l}),
\end{equation}
where the additive bias prevents complete suppression of RGB features, forcing methane cues to act as an enhancement signal rather than a replacement.

\subsubsection{\textbf{Semantic decoding and output head}}

The fused multi-scale features $ \{F_{1},\dots,F_{4}\} $ are decoded using a SegFormer \cite{xie2021segformer} decoder, producing dense semantic representation. As illustrated in Fig.~\ref{fig:Pipeline}, a lightweight refinement is applied after decoding. This refinement head consists of a small stack of convolutional blocks combining 3x3 convolutions, Group Normalization (GN), and Gaussian Error Linear Unit (GELU)~\cite{hendrycks2016gaussian} activations, which operate in feature space to locally smooth predictions and sharpen plume boundaries. Finally, a $1\times1$ convolution generates class logits, which are bilinearly upsampled to the original spatial resolution to obtain the final methane plume segmentation map. 

\subsubsection{\textbf{Loss function}}

The network is optimized end-to-end using a composite segmentation loss that combines a class-weighted focal loss~\cite{lin2017focal} and a Dice loss~\cite{milletari2016v}.

\vspace{2pt}
\noindent\textit{Dice loss.} We employ the Dice loss to directly optimize the overlap between the predicted mask and the ground truth:
\begin{equation}
\mathcal{L}_{\mathrm{Dice}}
=
1 -
\frac{
2 \sum_i y_i \hat{y}_i + \varepsilon
}{
\sum_i y_i + \sum_i \hat{y}_i + \varepsilon
},
\end{equation}
where $i$ indexes pixel locations, $y_i$ denotes the ground-truth label, $\hat{y}_i$ is the predicted probability, and $\varepsilon$ is a small constant for numerical stability.

\vspace{2pt}
\noindent\textit{Focal loss.} To mitigate class imbalance and emphasize hard-to-classify pixels, we adopt the focal loss formulation:
\begin{equation}
\mathcal{L}_{\mathrm{Focal}}
=
-
\sum_{i}
(1 - \hat{y}_i)^{\gamma}
\, y_i \log(\hat{y}_i),
\end{equation}
where $\gamma > 0$ is the focusing parameter that down-weights easy examples and encourages the model to focus on misclassified pixels.

\vspace{2pt}
\noindent\textit{Total loss.} The final training objective is defined as the sum of both terms:
\begin{equation}
\mathcal{L}
=
\mathcal{L}_{\mathrm{Dice}}
+
\mathcal{L}_{\mathrm{Focal}}.
\end{equation}
\section{Experiments and Results}

\subsection{Dataset}


To compare our approach with the state-of-the-art,
we use the EMIT methane plume dataset (MPDataset), introduced by Chen \emph{et al.}~\cite{mpsunet}. This dataset is derived from NASA’s Earth Surface Mineral Dust Source Investigation (EMIT) mission, available on the NASA Earthdata website \cite{NASA_EarthdataSearch}. using L1B At-Sensor Calibrated Radiance (EMITL1BRAD) for RGB imagery, L2B Methane Enhancement (EMITL2BCH4ENH) for methane plume enhancements and L2B Estimated Methane Plume Complexes data (EMITL2BCH4PLM). The methane enhancement maps are expressed in units of (ppm·m) corresponding to methane concentration (ppm) integrated along an atmospheric path length (m).

In the original MPSUNet study~\cite{mpsunet}, the dataset of 4,172 samples was divided into 3,004 training images, 334 validation images, and 834 test images. Although the original study reports the split sizes, the exact assignment of samples per subset is not publicly available. To ensure consistency with prior work, we preserve the same number of samples in each subset while adopting a unified and fixed data partition. The source code and the dataset split used in this work will be publicly released. 

The dataset covers the period from August 9, 2022,
to April 19, 2024. It consists of 4,172 samples with a spatial resolution of 512 × 512 pixels (approx. 60m/pixel). To balance positive and negative samples, the dataset applies data augmentation through 180-degree rotations.

\subsection{Experimental Setup}



All experiments were implemented in PyTorch and executed on a single NVIDIA RTX PRO 6000 Blackwell GPU. Unless otherwise specified, a fixed random seed of 42 was used. All modules of the proposed architecture were trained 
end-to-end without freezing any layers.

Our model was trained for 150 epochs using the AdamW optimizer with a learning rate of $6 \times 10^{-5}$, weight decay $5 \times 10^{-5}$, and a batch size of 16. The training objective combined Dice and Focal losses with equal weighting, and class weights were automatically estimated from the training data.

For all methods, including our model and the VGG16+UNet and SegFormer baselines, hyperparameters were optimized via Bayesian optimization on the validation set. VGG16+UNet and SegFormer were trained for 150 epochs too. MPSUNet was trained on the same data splits following the original training protocol, with the sole exception of the batch size, which was adjusted from 4 to 32 for computational considerations. All other hyperparameters, including 500 training epochs and the original random seed, were kept identical to the original configuration.

\subsection{Quantitative Results}

Quantitative results on the MPDataset are reported in Table~\ref{tab:comparison}. 
All metrics correspond to performance on the \textit{test split}. 

All reported metrics are computed exclusively on unseen test data. Performance is assessed using mean intersection over union $(MIoU = \frac{1}{N}\sum_{k=1}^{N}\frac{TP_{k}}{TP_{k} + FP_{k} + FN_{k}})$, mean precision $(MPrecision = \frac{1}{N}\sum_{k=1}^{N}\frac{TP_{k}}{TP_{k} + FP_{k}})$, recall for the plume class $(Recall = \frac{TP_{k}}{TP_{k} + FN_{k}})$, and mean pixel accuracy $(MPA = \frac{1}{N}\sum_{k=1}^{N}\frac{TP_{k}}{TP_{k} + FN_{k}})$, following the evaluation protocol of MPSUNet.

As shown in Table~\ref{tab:comparison}, the proposed method consistently outperforms state-of-the-art models across all reported metrics. Compared to MPSUNet, our approach achieves improvements of $+0.92$ in MIoU, $+0.87$ in MPrecision, $+1.01$ in Recall, and $+0.52$ in MPA. While SegFormer is the most lightweight model and the fastest in terms of inference time, the proposed approach requires approximately 80\% fewer FLOPs than MPSUNet and VGG16+UNet, while maintaining a competitive inference time (IT) of 7.06 ms, representing the average per-sample inference time over multiple runs. Despite containing more parameters, the proposed model is computationally more efficient than previous high-performing baselines, benefiting from the transformer architecture efficiency and lower-resolution feature processing.

\begin{figure*}[!t]
\centering
\label{tab:comparisson_qualitative}

\setlength{\tabcolsep}{2pt}
\renewcommand{\arraystretch}{1}

\begin{tabular}{c c c c c c c c}
\toprule
\textbf{} 
& \textbf{RGB} 
& \textbf{ENH} 
& \textbf{GT} 
& \textbf{VGG16+UNet} 
& \textbf{SegFormer} 
& \textbf{MPSUNet} 
& \textbf{Ours} \\
\midrule

\raisebox{1.5\height}{\rotatebox[origin=c]{90}{\textbf{Scene 1}}}&
\includegraphics[width=0.12\textwidth]{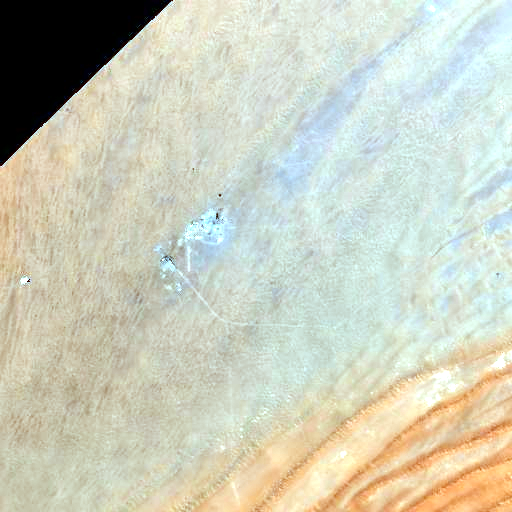} &
\includegraphics[width=0.12\textwidth]{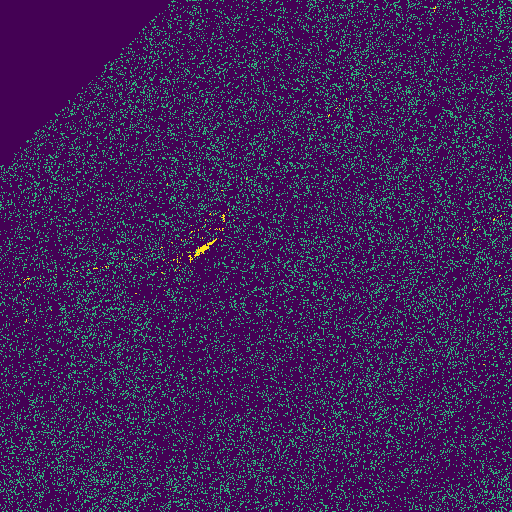} &
\includegraphics[width=0.12\textwidth]{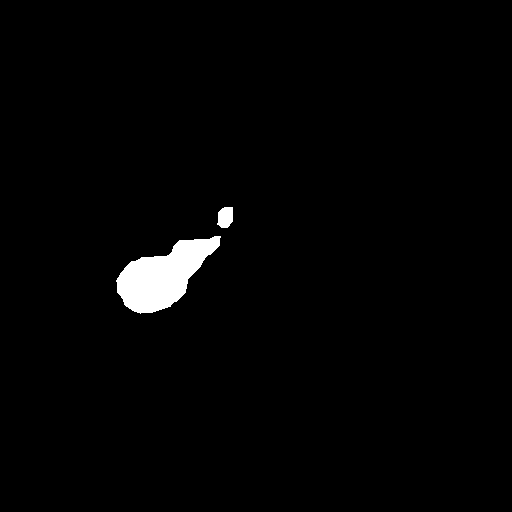} &
\includegraphics[width=0.12\textwidth]{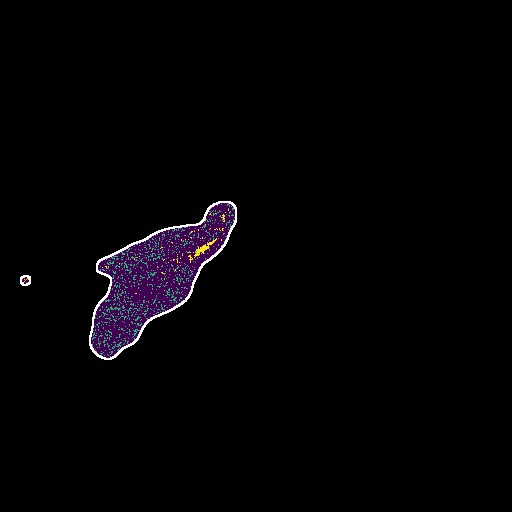} &
\includegraphics[width=0.12\textwidth]{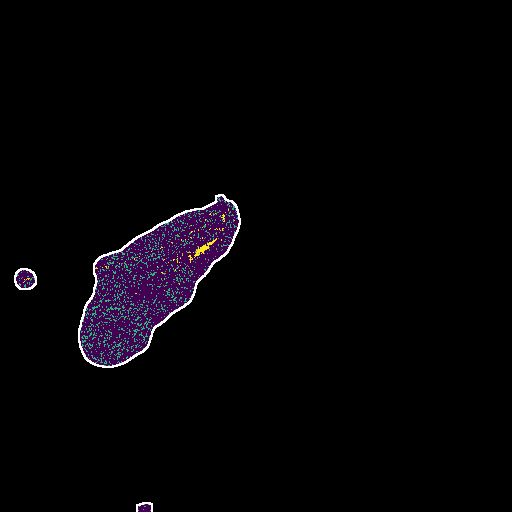} &
\includegraphics[width=0.12\textwidth]{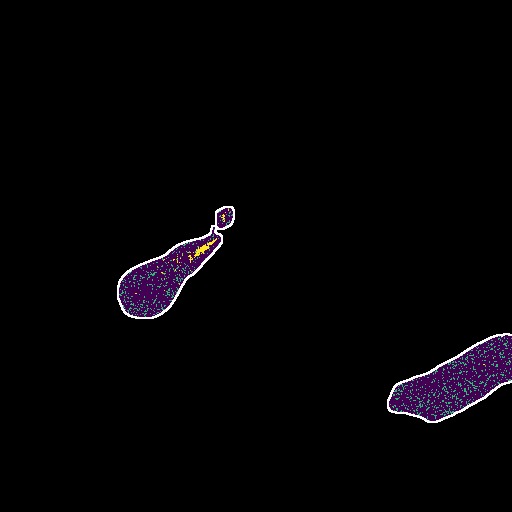} &
\includegraphics[width=0.12\textwidth]{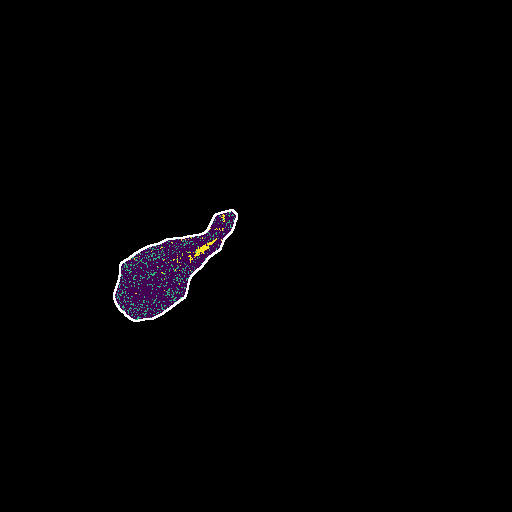} \\

\raisebox{1.5\height}{\rotatebox[origin=c]{90}{\textbf{Scene 2}}}&
\includegraphics[width=0.12\textwidth]{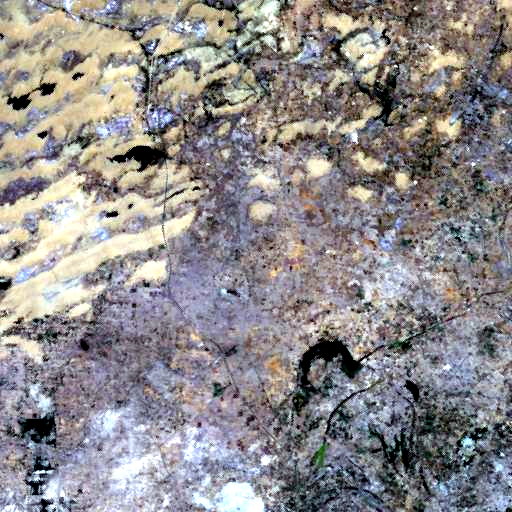} &
\includegraphics[width=0.12\textwidth]{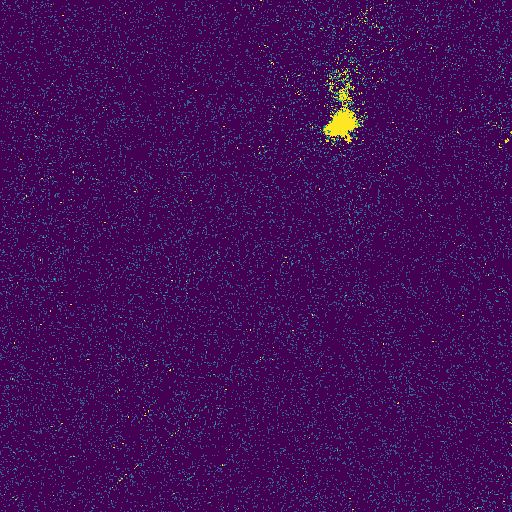} &
\includegraphics[width=0.12\textwidth]{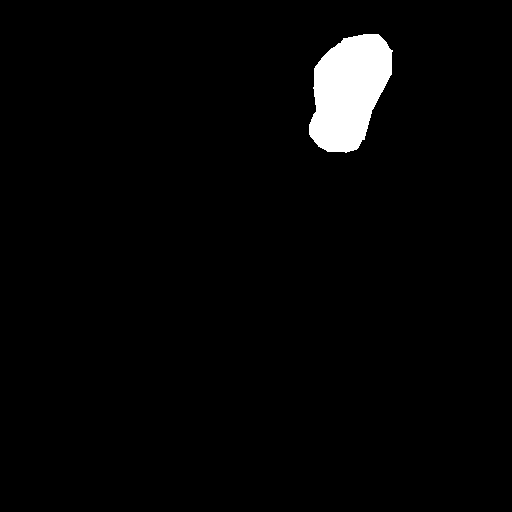} &
\includegraphics[width=0.12\textwidth]{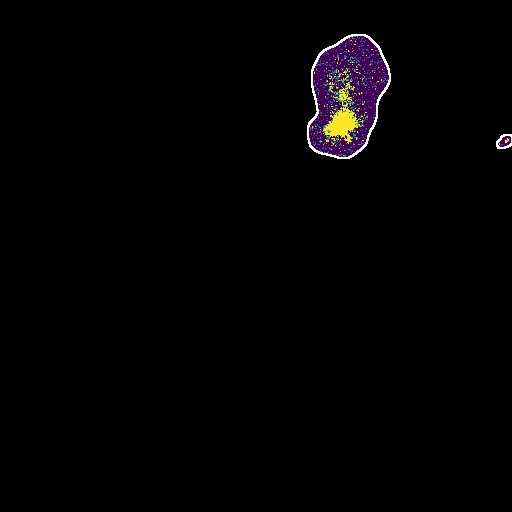} &
\includegraphics[width=0.12\textwidth]{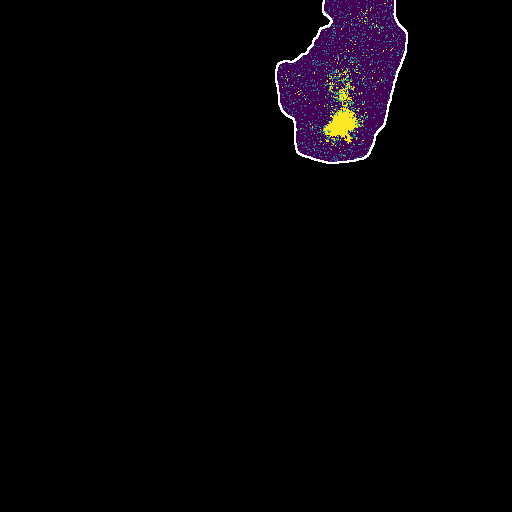} &
\includegraphics[width=0.12\textwidth]{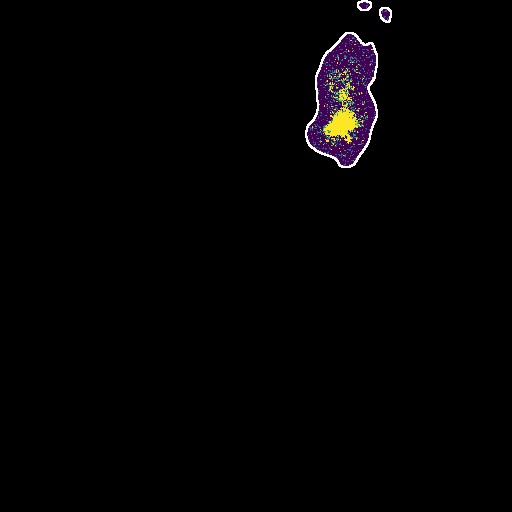} &
\includegraphics[width=0.12\textwidth]{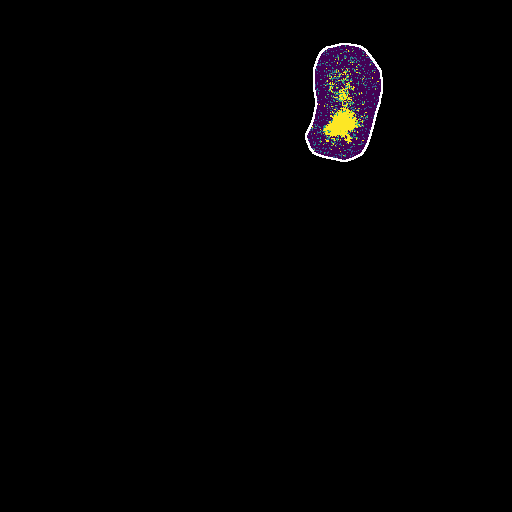} \\

\raisebox{1.5\height}{\rotatebox[origin=c]{90}{\textbf{Scene 3}}}&
\includegraphics[width=0.12\textwidth]{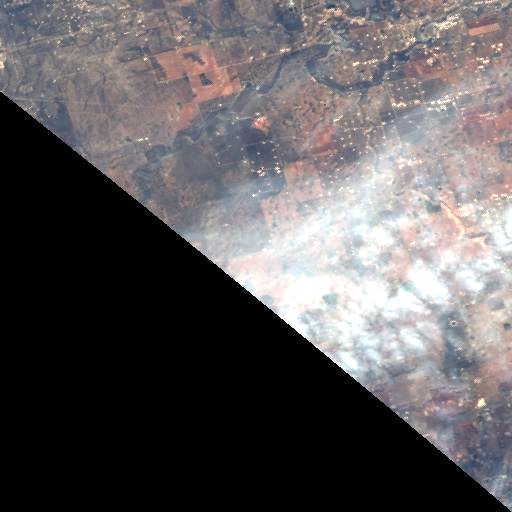} &
\includegraphics[width=0.12\textwidth]{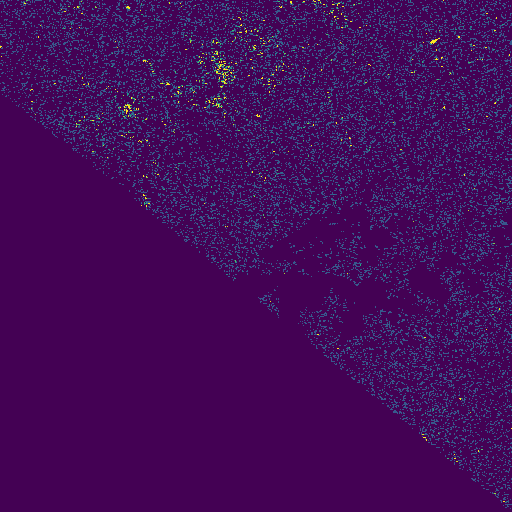} &
\includegraphics[width=0.12\textwidth]{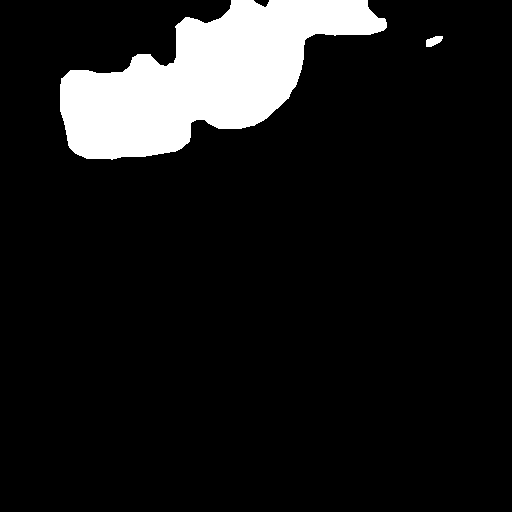} &
\includegraphics[width=0.12\textwidth]{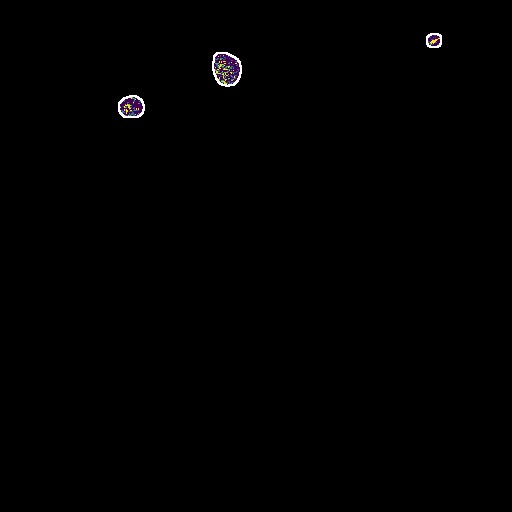} &
\includegraphics[width=0.12\textwidth]{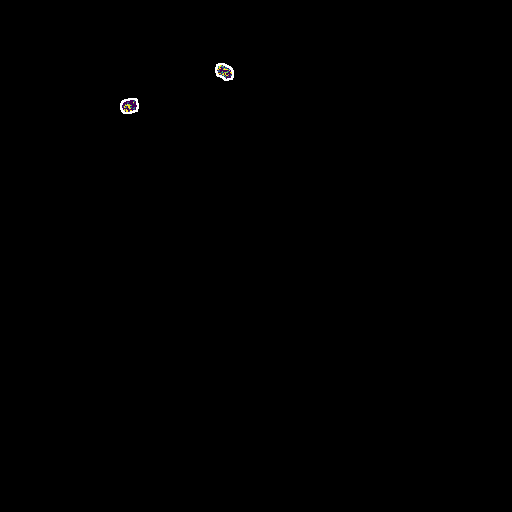} &
\includegraphics[width=0.12\textwidth]{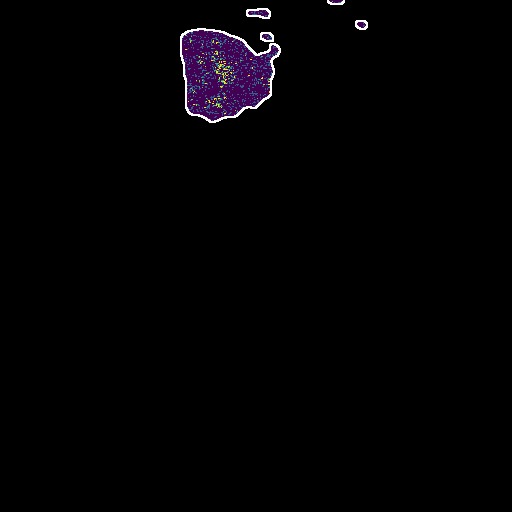} &
\includegraphics[width=0.12\textwidth]{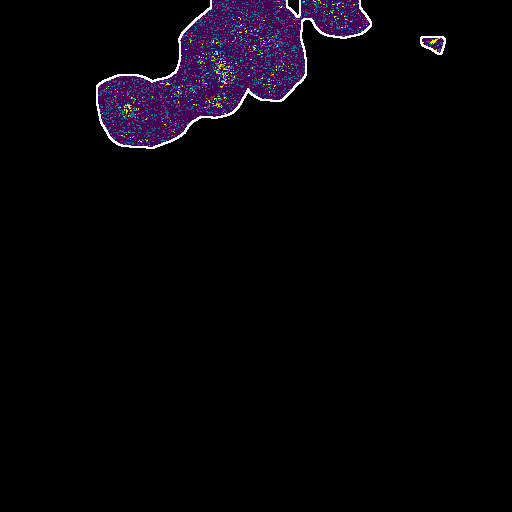} \\

\bottomrule
\end{tabular}

\caption{Qualitative comparison of methane plume segmentation results on the MPDataset. 
Columns show RGB input, methane enhancement (ENH), ground truth image (GT), and predictions from VGG16+UNet, SegFormer, MPSUNet, and our proposed method.}
\label{fig:qualitative_comparison}
\end{figure*}

\begin{table}[!t]
\raggedright
\caption{\MakeUppercase{Quantitative comparison of methane plume segmentation models}}
\label{tab:comparison}
\renewcommand{\arraystretch}{1.2}
\setlength{\tabcolsep}{5pt}
{\fontsize{13}{15}\selectfont
\resizebox{\columnwidth}{!}{%
\begin{tabular}{lccccccc}
\toprule
Model & MIoU (\%) & MPrecision (\%) & Recall (\%) & MPA (\%) & FLOPs (G) & Params (M) & IT (ms) \\
\midrule
VGG16+UNet  & 76.74 & 84.92 & 70.31 & 84.99 & 225.98 & 24.89 &  6.74  \\
SegFormer   & 73.11 & 83.86 & 60.19 & 79.94 & \textbf{6.79} & \textbf{3.71} & \textbf{2.42}\\
MPSUNet \cite{mpsunet}    & 81.23 & 89.20 & 75.92 & 87.84 & 226.05 & 24.93 & 15.78 \\
\textbf{Ours} & \textbf{82.15} & \textbf{90.07} & \textbf{76.93} & \textbf{88.36} & 34.75 & 36.78 & 7.06 \\
\bottomrule
\end{tabular}
}%
}
\end{table}

\subsection{Qualitative Results}


Figure~\ref{fig:qualitative_comparison} presents a qualitative comparison of methane plume segmentation results across three representative scenes, highlighting the different behaviors of the evaluated models under varying plume characteristics and background complexity.

In Scene~1, the plume is subtle and poorly contrasted, making localization particularly challenging. VGG16+UNet and SegFormer partially identify the plume core but produce fragmented predictions with limited spatial accuracy. MPSUNet fails to correctly localize the plume, missing the main plume structure entirely, which highlights its reduced sensitivity under low-contrast enhancement conditions. In contrast, the proposed method successfully identifies the plume region and produces a compact and well-localized segmentation that aligns with the enhancement map and ground truth.

In Scene~2, the methane plume is more clearly distinguishable in the enhancement map, enabling improved detection across models. VGG16+UNet produces a strong segmentation of the main plume region but introduces a small spurious plume at the boundaries, indicating localized false positives. A similar behavior is observed for MPSUNet, which correctly captures the main plume but also predicts small artificial plume fragments near the plume edges. SegFormer again exhibits a tendency toward over-segmentation, generating plume regions not supported by the enhancement map. In contrast, the proposed method accurately segments the plume without introducing spurious detections, maintaining clean boundaries and strong consistency with both the enhancement map and ground truth.

In Scene~3, the plume exhibits a diffuse spatial extent with varying intensity levels. VGG16+UNet and SegFormer largely fail to capture the plume structure, detecting only small isolated regions. MPSUNet identifies multiple plume fragments, including several spurious detections in regions with no methane enhancement, effectively hallucinating plume structures in low-confidence areas. In contrast, the proposed approach focuses on the high-confidence plume core, producing a smaller but spatially coherent segmentation that matches the most reliable enhancement regions. This behavior reflects the precision-oriented trade-off observed in the quantitative results, favoring reliable plume delineation over exhaustive coverage of low-intensity or uncertain plume extensions.

Overall, these qualitative examples corroborate the quantitative analysis in Table~\ref{tab:comparison}, illustrating complementary segmentation characteristics across models and demonstrating that the proposed method consistently prioritizes accurate plume localization, boundary precision, and robustness against false positives across diverse scenarios.

\section{Conclusions and Future Work}

In this work, we proposed a multimodal deep learning framework for methane plume segmentation that integrates a feature-guided methane enhancement (FGME) mechanism to inject physically meaningful methane cues into RGB transformer representations across multiple semantic scales. By effectively coupling spectral methane information with high-level visual features, the proposed approach improves plume discrimination while maintaining architectural efficiency.

The experimental results show that the proposed method consistently outperforms all evaluated baselines. In addition to achieving the best overall segmentation performance, our model provides a favorable trade-off between accuracy and computational efficiency. As future work, we plan to extend the current framework by incorporating additional spectral information from wavelength regions with strong methane absorption, enabling a more physically grounded representation of gas signatures.

\section{Acknowledgments}

This work was supported by the Air Force Office of Scientific Research (AFOSR) through the Southern Office of Aerospace Research and Development (SOARD) under grant number FA8655-25-1-8010.


\small
\bibliographystyle{IEEEtranN}
\bibliography{references}

\end{document}